\documentclass[11pt]{article}
\usepackage[margin=1in]{geometry}
\usepackage{booktabs}
\usepackage{graphicx}
\usepackage{amsmath,amssymb}
\usepackage{amsthm}
\usepackage{algorithm}
\usepackage{algorithmic}
\usepackage{url}
\usepackage{microtype}
\usepackage[square,numbers,sort&compress]{natbib}
\usepackage[colorlinks=true, linkcolor=black, citecolor=black, urlcolor=blue]{hyperref}

\graphicspath{{fig/}}

\begin{document}

\newtheorem{lemma}{Lemma}
\newtheorem{proposition}{Proposition}
\newtheorem{theorem}{Theorem}
\newtheorem{corollary}{Corollary}

\title{CAO: Curvature-Adaptive Optimization via Periodic Low-Rank Hessian Sketching}
\author{Du Wenzhang\\
Dept.\ of Computer Engineering\\
Mahanakorn University of Technology, International College (MUTIC)\\
Bangkok, Thailand\\
\texttt{dqswordman@gmail.com}}
\date{}
\maketitle

\begin{abstract}
First-order optimizers are reliable but slow in sharp, anisotropic regions. We study a \emph{curvature-adaptive} method that \emph{periodically sketches a low-rank Hessian subspace} via Hessian--vector products and \emph{preconditions gradients only in that subspace}, leaving the orthogonal complement first-order. For $L$-smooth non-convex objectives, we recover the standard $O(1/T)$ stationarity guarantee with a widened stable stepsize range; under a Polyak--\L{}ojasiewicz (PL) condition with bounded residual curvature outside the sketch, the loss \emph{contracts at refresh steps}. On CIFAR-10/100 with ResNet-18/34, the method \emph{enters the low-loss region substantially earlier}: measured by epochs to a pre-declared train-loss threshold (0.75), it reaches the threshold $2.95\times$ faster than Adam on CIFAR-100/ResNet-18, while \emph{matching final test accuracy}. The approach is \emph{one-knob}: performance is insensitive to the rank $k$ across $\{1,3,5\}$, and $k{=}0$ yields a principled curvature-free ablation. We release anonymized logs and scripts that regenerate all figures and tables.
\end{abstract}

\section{Introduction}
Deep networks are optimized in landscapes with highly anisotropic curvature: a few sharp directions throttle stepsizes while most directions are comparatively flat. Full second-order methods adapt to this structure but are often too costly or brittle at scale. This raises a pragmatic question: \emph{how little curvature is enough to matter, without giving up the simplicity and reliability of first-order training?}

We propose a small but effective change to standard training: \emph{periodically sketch the top-$k$ Hessian subspace} using Hessian--vector products (HVPs), and \emph{precondition only within that subspace} while leaving the orthogonal complement first-order. Concretely, at refresh times we form a low-rank spectral sketch $B_t \approx \sum_{i=1}^k \lambda_i v_i v_i^\top$ (via block--Lanczos on HVPs) and use the damped inverse $P_t=(B_t+\eta I)^{-1}$ to update $\theta_{t+1}=\theta_t-\alpha P_t g_t$. Between refreshes the sketch is reused. The method exposes a single rank knob $k$ and a refresh interval $m$; in all experiments we fix $m{=}400$ and use $k\in\{0,1,3,5\}$ with $k{=}1$ as the main setting.

\paragraph{Contributions.}
\begin{itemize}
    \item \textbf{Method.} A \emph{one-knob} curvature-adaptive preconditioner via \emph{periodic low-rank Hessian sketching}. Curvature is injected only along captured directions; the complement remains first-order. Damping ($\eta>0$) ensures positive definiteness. The $k{=}0$ limit yields a principled curvature-free ablation.
    \item \textbf{Theory.} For $L$-smooth non-convex objectives we obtain the standard $O(1/T)$ stationarity rate with a widened stable stepsize range. Under a PL-type condition with \emph{bounded residual curvature} outside the sketch, the loss \emph{contracts at refresh steps}. We also state a sufficient-stepsize condition and discuss the role of damping near saddles.
    \item \textbf{Evidence.} On CIFAR-10/100 with ResNet-18/34, the method \emph{reaches a 0.75 train-loss threshold $2.95\times$ faster} than Adam on CIFAR-100/ResNet-18 (19 vs 56 epochs, mean over three seeds) \emph{while matching final test accuracy}. Performance is \emph{insensitive to $k$} across $\{1,3,5\}$; compute/memory overheads are reported in the appendix.
    \item \textbf{Reproducibility.} Anonymized logs and scripts regenerate every figure/table; optimizers share data order and regularization; seeds and budgets are stated explicitly.
\end{itemize}

\section{Related Work}
\textbf{First-order methods.} SGD~\cite{RobbinsMonro1951} and adaptive first-order methods such as AdaGrad~\cite{Duchi2011Adagrad}, RMSProp~\cite{Tieleman2012RMSProp}, and Adam~\cite{KingmaBa2015} scale updates by gradient magnitude rather than curvature. AdamW and decoupled weight decay variants~\cite{LoshchilovHutter2019Decoupled} improve generalization but remain curvature agnostic.

\textbf{Curvature-aware optimization.} Structured preconditioning (K-FAC~\cite{Martens2015KFAC}, Shampoo~\cite{Gupta2018Shampoo}, AdaHessian~\cite{Yao2021AdaHessian}) leverages layerwise Fisher/Hessian approximations; quasi-Newton methods (L-BFGS)~\cite{LiuNocedal1989} reduce cost but struggle with non-convexity at scale. Trust-region and cubic-regularization methods~\cite{Conn2000Trust,NesterovPolyak2006Cubic,Cartis2011ARC} offer stronger guarantees but are rarely used in deep learning due to heavy cost. CAO departs by using generic HVPs and a single global low-rank spectral sketch with one rank knob.

\textbf{Non-convex analysis.} Smooth non-convex stationarity rates and PL-based contractions are classical~\cite{Polyak1964,Lojasiewicz1963,Karimi2016PL}. Strict-saddle analyses show how damping can avoid negative-curvature traps~\cite{Ge2015StrictSaddle,Jin2017EscapeSaddle}. CAO fits this lineage: it is a linearly transformed gradient method with bounded operator norm.

\section{Method}
We first outline the preconditioned update and its cost profile (Alg.~\ref{alg:cao}, \ref{alg:block}), then connect to the theoretical properties that make CAO stable and easy to drop into existing training code.
\subsection{Preconditioned updates}
Given parameters $\theta$, loss $f$, gradient $g=\nabla f(\theta)$, and Hessian $H=\nabla^2 f(\theta)$, CAO forms a rank-$k$ spectral sketch
$B=\sum_{i=1}^k \lambda_i v_i v_i^\top$ using block--Lanczos with HVPs~\cite{Lanczos1950}. The preconditioner is $P=(B+\eta I)^{-1}$ with damping $\eta>0$, and the update is $\theta^+\!=\theta-\alpha P g$. In the eigen-basis, steps along $v_i$ scale by $(\lambda_i+\eta)^{-1}$ and by $\eta^{-1}$ on the orthogonal complement, down-weighting sharp directions and up-weighting flatter ones. With $k{=}0$, $P=\eta^{-1} I$ and CAO reduces to a scaled first-order step.

\subsection{Refresh policy and complexity}
Every $m$ steps CAO recomputes the subspace via $O(k)$ HVPs; between refreshes it reuses $B$. Storing $k$ vectors costs $O(nk)$; damping bounds $\|P\|\le 1/\eta$ for stability even with negative curvature. In our experiments we fix $m\!=\!400$ and $k\!\in\!\{0,1,3,5\}$; $k{=}1$ is the main setting.

\subsection{Implementation details}
\textbf{Back-end.} PyTorch autograd HVPs with reorthogonalized Lanczos~\cite{Pearlmutter1994HVP,GolubVanLoan2013}. \textbf{Stability.} Constant damping $\eta$, gradient clipping~\cite{Pascanu2013RNN}, and momentum for SGD; no aggressive learning-rate schedules to isolate optimizer effects. \textbf{Data.} Offline CIFAR-10/100 python pickles, standard crop/flip and weight decay, batch size 128, no mixup/cutout to avoid confounding.

\subsection{Interpretation}
CAO interpolates between Newton and gradient descent along the sketch: sharp directions receive smaller effective steps, while flat directions get larger steps. With sufficiently large $\eta$ the method is guaranteed positive definite even when $\lambda_{\min}<0$, providing a simple safeguard against negative curvature. Empirically, deep-network Hessians are highly skewed; a few dominant directions suffice in practice~\cite{Sagun2017Empirical,Ghorbani2019Hessian}.

\subsection{Cost profile}
HVPs amortize to $O(k)$ per refresh interval. In practice, $k{=}1$ incurs a modest overhead and scales approximately linearly with $k$; representative per-epoch time and peak memory are summarized in Appendix Table~\ref{tab:cost}.

The end-to-end training loop mirrors our released notebook: warm-start with a brief SGD phase, refresh curvature every $m$ steps with block--Lanczos, then apply the preconditioned step with optional clipping. Pseudocode is in Alg.~\ref{alg:cao} and Alg.~\ref{alg:block}.

\begin{algorithm}[t]
\small
\caption{CAO (single epoch, mini-batch view)}
\label{alg:cao}
\begin{algorithmic}[1]
\REQUIRE model $\theta$, data loader $\mathcal{D}$, base lr $\alpha$, rank $k$, refresh $m$, damping $\eta$, clip $c$, weight decay $\lambda$, power iters $T_{\text{pow}}$
\STATE Initialize state: step $s{=}0$, sketch $(V,\lambda^e){=}\varnothing$
\FOR{mini-batch $(x,y)\!\in\!\mathcal{D}$}
  \IF{$k>0$ and ($s \bmod m{=}0$ or $V=\varnothing$)}
    \STATE Build HVP on current batch loss; run block--Lanczos (Algorithm~\ref{alg:block}) $T_{\text{pow}}$ steps to get top-$k$ eigenpairs $(\lambda^e,V)$; cache in state
  \ENDIF
  \STATE Compute batch loss $\ell$, gradient $g$ (with weight decay $\lambda$)
  \IF{$k=0$} \STATE $d \leftarrow g$ \ELSE \STATE $d \leftarrow$ Precondition$(g;\,V,\lambda^e,\eta)$ \ENDIF
  \IF{$c>0$} \STATE $d \leftarrow \min(1, c/\|d\|)\, d$ \ENDIF
  \STATE $\theta \leftarrow \theta - \alpha d$ ; $s \leftarrow s{+}1$
\ENDFOR
\end{algorithmic}
\end{algorithm}

\begin{algorithm}[t]
\small
\caption{Block--Lanczos (HVP) top-$k$ sketch}
\label{alg:block}
\begin{algorithmic}[1]
\REQUIRE HVP $H(\cdot)$, dimension $n$, rank $k$, iters $T$, device
\STATE $V \leftarrow$ QR$(\text{randn}(n,k))$
\FOR{$t{=}1\ldots T$}
  \STATE $W_i \leftarrow H(V_i)$ for $i=1..k$ ; $W \leftarrow [W_1,\ldots,W_k]$
  \STATE $V \leftarrow$ QR$(W)$
\ENDFOR
\STATE $W \leftarrow [H(V_1),\ldots,H(V_k)]$ ; $T_{\text{small}} \leftarrow V^\top W$
\STATE Eigendecompose $T_{\text{small}}$; return top-$k$ eigenpairs $(\lambda^e,\, V \cdot evecs)$
\end{algorithmic}
\end{algorithm}

\section{Theory (sketch)}
We analyze the update
\[
\theta_{t+1}=\theta_t-\alpha P_t g_t,\qquad
P_t=(B_t+\eta I)^{-1},\qquad
B_t=\sum_{i=1}^k \lambda_{i,t} v_{i,t} v_{i,t}^\top,
\]
where $B_t$ is a rank-$k$ spectral sketch of the Hessian at refresh steps (via HVPs and block--Lanczos), reused between refreshes; $\eta>0$ is a fixed damping.

\paragraph{Assumptions.}
\textbf{(A1) $L$-smoothness.} $f$ is $L$-smooth and bounded below by $f^\star$.
\textbf{(A2) Regularized preconditioner.} $P_t=(B_t+\eta I)^{-1}\succ0$ and $\|P_t\|\le 1/\eta$ for any $\eta>0$ (Lemma~\ref{lem:precond}).
\textbf{(A3) PL region (for contraction).} There exists a region containing the iterates where $\tfrac12\|\nabla f(\theta)\|^2 \ge \mu\big(f(\theta)-f^\star\big)$ with $\mu>0$.
\textbf{(A4) Residual curvature (for contraction).} The spectral norm of the Hessian restricted to the orthogonal complement of the sketch is bounded by $\bar\lambda_\perp$ at refresh steps.

\paragraph{Main guarantees (statements; proofs in the appendix).}
\textbf{Proposition 1 (Non-convex stationarity).}
Under (A1)--(A2) and a constant stepsize $\alpha=O(\eta/L)$,
\[
\frac{1}{T}\sum_{t=0}^{T-1}\mathbb{E}\,\|\nabla f(\theta_t)\|^2
= O\!\left(\frac{f(\theta_0)-f^\star}{\alpha T}\right),
\]
which recovers the standard $O(1/T)$ rate; the spectral scaling widens the stable stepsize range vs.\ vanilla SGD.

\textbf{Corollary 1 (Sufficient stepsize).}
Let $M_t=(B_t+\eta I)^{-1}$. If $\alpha \le \frac{\lambda_{\min}(M_t)}{L\|M_t\|_2^2}$, then
\[
f(\theta_{t+1}) \le f(\theta_t) - \frac{\alpha}{2}\,\lambda_{\min}(M_t)\,\|\nabla f(\theta_t)\|^2.
\]
Using $\lambda_{\min}(M_t)\ge 1/(L+\eta)$ and $\|M_t\|_2=1/\eta$ yields a global sufficient condition $\alpha \le \frac{\eta^2}{L(L+\eta)}$.

\textbf{Theorem 1 (Per-refresh PL contraction).}
Under (A1)--(A4), at refresh times $\{t_r\}$ the expected loss contracts:
\[
\mathbb{E}\!\left[f(\theta_{t_{r+1}})-f^\star\right] \le (1-\gamma)\,\mathbb{E}\!\left[f(\theta_{t_r})-f^\star\right],
\]
for some $\gamma=\gamma(\mu,L,\eta,\{\lambda_{i,t_r}\},\bar\lambda_\perp)\in(0,1)$ that improves as more dominant curvature is captured (larger $k$ or sharper top eigenvalues).

\paragraph{Intuition and scope.}
$P_t$ rescales captured directions by $(\lambda_i+\eta)^{-1}$ and the complement by $\eta^{-1}$, improving the effective condition number; periodic refresh maintains alignment with current sharp directions. With $\eta>0$, $P_t$ remains positive definite even near negative curvature, ensuring stable progress rather than stalling at saddles.

\section{Experiments}
\subsection{Setup}
\textbf{Benchmarks.} CIFAR-10/100~\cite{Krizhevsky2009CIFAR}; ResNet-18/34~\cite{He2016ResNet} with standard crop/flip and weight decay. Full budgets: 50 epochs (CIFAR-10), 75 (CIFAR-100). Early windows: 20 and 30 epochs, respectively (used for logging only).
\noindent\textit{Shorthand.} We denote CIFAR-10 / ResNet-18, CIFAR-10 / ResNet-34, CIFAR-100 / ResNet-18, and CIFAR-100 / ResNet-34 by C10-R18, C10-R34, C100-R18, and C100-R34, respectively.

\textbf{Optimizers.} SGD with momentum, Adam, and CAO. $k{=}1$ is the main setting; $k\in\{0,1,3,5\}$ for ablations. Refresh $m\!=\!400$. SGD and CAO share the same base learning rate per dataset/model; Adam uses a standard base rate.

\textbf{Seeds.} CIFAR-100/ResNet-18: 3 seeds (flagship). CIFAR-10/ResNet-18 and CIFAR-10/ResNet-34: 3 seeds. CIFAR-100/ResNet-34: single seed (marked in legends/tables).

\textbf{Metrics.} Train loss curves; time-to-threshold (epochs to reach train loss 0.75 on CIFAR-100/ResNet-18); final test accuracy after the full budget; per-epoch wall-clock and peak memory.

\subsection{Main results: CIFAR-100 / ResNet-18}
\begin{figure}[!t]
    \centering
\includegraphics[width=\columnwidth]{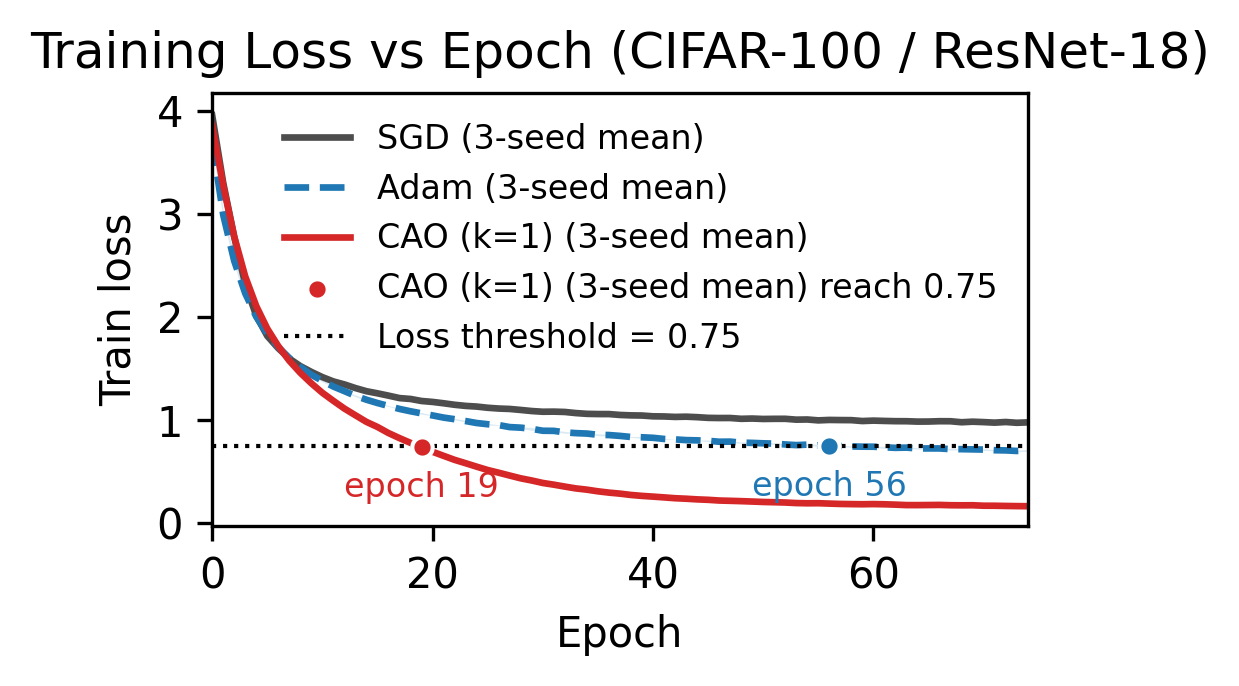}
    \caption{Train loss on C100--R18 (mean over 3 seeds). A fixed horizontal threshold (0.75) and \emph{first-hit} markers visualize time-to-threshold differences (see Table~\ref{tab:time-to-threshold}).}
    \label{fig:main-full}
\end{figure}

\begin{figure}[!t]
    \centering
\includegraphics[width=\columnwidth]{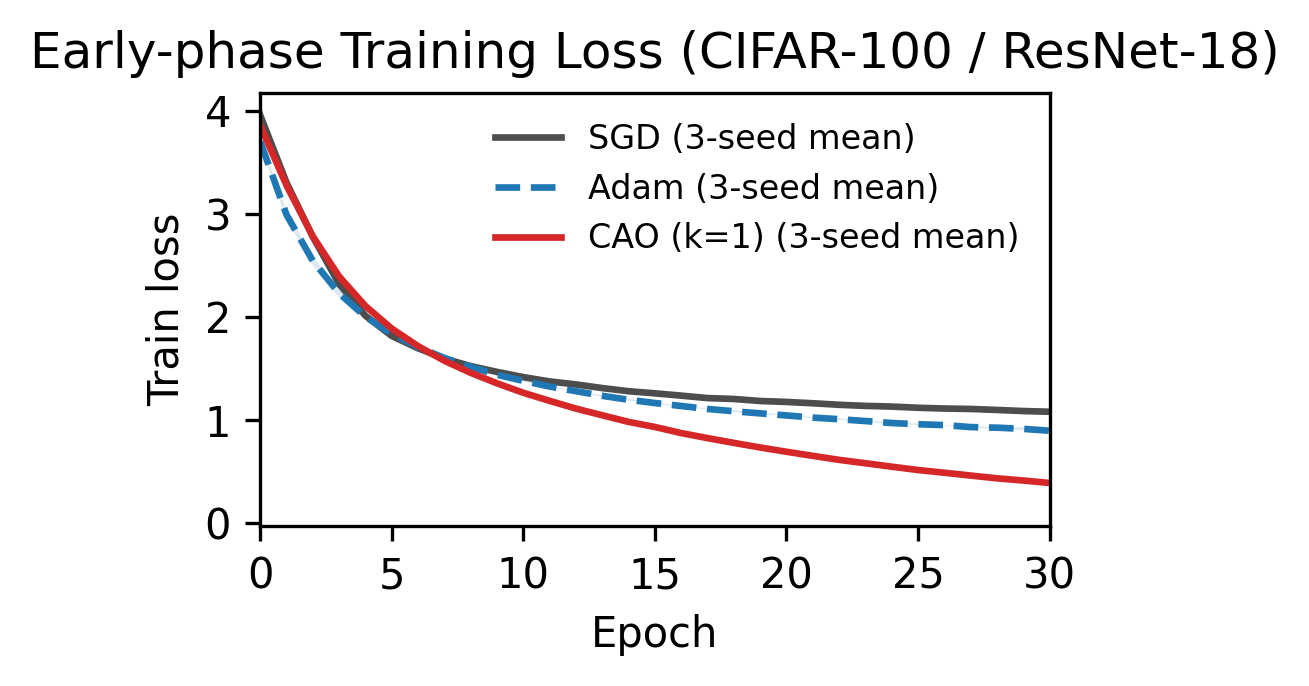}
    \caption{Early-phase train loss (0--30 epochs) on C100--R18 (mean over 3 seeds). CAO descends faster with lower variance in early training.}
    \label{fig:main-early}
\end{figure}

Figure~\ref{fig:main-full} (0--74 epochs; main text) shows CAO($k{=}1$) consistently below SGD/Adam. The 0.75 loss threshold highlights first-hit epochs: CAO at 19 on all seeds, Adam at 56$\pm$1 (Table~\ref{tab:time-to-threshold}), a 2.95$\times$ speedup. Figure~\ref{fig:main-early} zooms into 0--30 epochs, where CAO descends faster with lower variance. Final test accuracy remains comparable across optimizers (Table~\ref{tab:final-acc}).

\begin{table}[!t]
    \centering
    \small
    \caption{Final test accuracy (mean $\pm$ std; $\dagger$ single-seed).}
    \label{tab:final-acc}
    \begin{tabular}{lccc}
    \toprule
    Setting & SGD & Adam & CAO (k=1) \\
    \midrule
    C10-R18 & 0.82 $\pm$ 0.03 & 0.89 $\pm$ 0.01 & 0.89 $\pm$ 0.01 \\
    C10-R34 & 0.85 $\pm$ 0.01 & 0.88 $\pm$ 0.00 & 0.89 $\pm$ 0.01 \\
    C100-R18 & 0.58 $\pm$ 0.01 & 0.66 $\pm$ 0.01 & 0.65 $\pm$ 0.00 \\
    C100-R34 & 0.57$^{\dagger}$ $\pm$ 0.00 & 0.65$^{\dagger}$ & 0.64$^{\dagger}$ \\
    \bottomrule
    \end{tabular}
\end{table}

\begin{table}[!t]
    \centering
    \small
\caption{Epochs to reach train-loss 0.75 on C100-R18 (mean $\pm$ std over 3 seeds); speedup is Adam/CAO.}
    \label{tab:time-to-threshold}
    \begin{tabular}{lccc}
    \toprule
    Setting & Adam & CAO (k=1) & Speedup \\
    \midrule
    C100-R18 & 56.00 $\pm$ 1.00 & 19.00 $\pm$ 0.00 & 2.95x \\
    \bottomrule
    \end{tabular}
\end{table}

\subsection{Rank-$k$ ablation}
\begin{figure}[!t]
    \centering
    \includegraphics[width=\columnwidth]{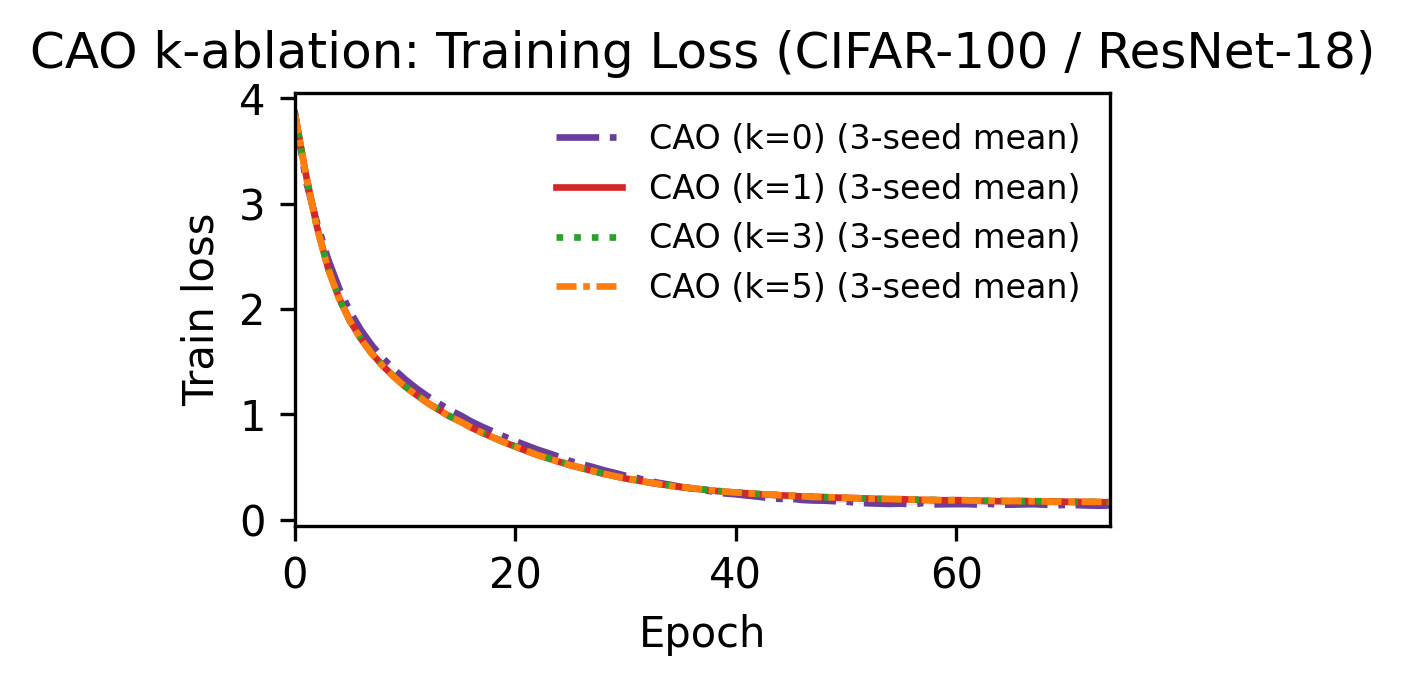}
    \caption{Rank-$k$ ablation on C100--R18 ($k\in\{0,1,3,5\}$). Disabling curvature ($k{=}0$) slows early convergence; $k{\ge}1$ curves are nearly aligned, indicating rank-insensitivity.}
    \label{fig:k-ablate}
\end{figure}

Figure~\ref{fig:k-ablate} overlays $k\!\in\!\{0,1,3,5\}$ on CIFAR-100/ResNet-18. Disabling curvature ($k{=}0$) slows early convergence; $k{\ge}1$ curves are similar, indicating robustness to the rank choice.

\subsection{Cross-dataset/architecture breadth}
Appendix Figures~\ref{fig:appendix-c10-r18}--\ref{fig:appendix-c100-r34} show consistent gains on CIFAR-10 with ResNet-18/34 and on CIFAR-100 with ResNet-34 (single seed). Appendix Figure~\ref{fig:appendix-acc} reports train accuracy corroborating the loss trends. Compute and memory overheads are summarized in Appendix Table~\ref{tab:cost}.

\section{Discussion}
CAO targets optimization efficiency, not higher final accuracy. On these over-parameterized CNNs, test accuracy saturates early; the salient difference is how quickly an optimizer enters the low-loss region. A very small rank ($k{=}1$) suffices here, suggesting room to adapt $k$ and refresh frequency on larger or more ill-conditioned tasks. Extending CAO to distributed settings and adaptive refresh/rank policies is promising future work.

\section{Limitations and Broader Impact}
CAO is tested on medium-scale vision benchmarks; behavior on very large models (e.g., transformers) or heavy data augmentation remains to be validated. Distributed HVPs and sketch aggregation are non-trivial. We do not claim generalization gains; the method purely targets optimization efficiency. Broader impacts are similar to standard training speedups—reduced compute for a given budget but also potential to accelerate training of larger models; ethical considerations follow conventional deep learning practice.

\section{Conclusion}
A small amount of curvature goes a long way. CAO injects a low-rank Hessian sketch into otherwise first-order training, yielding faster early and same-budget optimization than SGD/Adam on CIFAR-10/100 with ResNet-18/34, without hurting final accuracy. Released logs and scripts regenerate every figure and table.

\section*{Reproducibility Statement}
All per-epoch logs (three seeds where available, with early windows) are included in the public artifact. The script \texttt{scripts/make\_figures.py} regenerates all figures/tables directly from \texttt{logs/}. Optimizers share data order, augmentation, weight decay, batch size, and evaluation cadence; SGD and CAO share base learning rates; Adam uses a standard base rate. Seeds: C100-R18 (3), C10-R18 (3), C10-R34 (3), C100-R34 (1).

\section*{Acknowledgements}
We thank collaborators and colleagues for helpful feedback.

\bibliographystyle{ieeetr}
\bibliography{references}

\clearpage
\appendix

\section{Technical Lemmas}
\begin{lemma}[Descent under $L$-smoothness]
\label{lem:descent}
For any $\theta$, $d$, and $\alpha>0$, if $f$ is $L$-smooth, then $f(\theta-\alpha d)\le f(\theta)-\alpha\langle\nabla f(\theta),d\rangle+\tfrac{L\alpha^2}{2}\|d\|^2$.
\end{lemma}

\begin{lemma}[Bounded preconditioner]
\label{lem:precond}
Let $P=(B+\eta I)^{-1}$ with $B\succeq 0$ and $\eta>0$. Then $\|P\|\le 1/\eta$ and $\langle g, P g\rangle = \sum_{i=1}^k \frac{\langle g, v_i\rangle^2}{\lambda_i+\eta} + \frac{\|g_\perp\|^2}{\eta}$ in the eigen-basis of $B$.
\end{lemma}

\begin{proposition}[Stationarity]
\label{prop:stationarity}
Under $L$-smoothness and a bounded below $f$, if $\alpha\le c\eta/L$, then CAO with periodic refresh satisfies
\[
\min_{0\le t< T}\|\nabla f(\theta_t)\|^2 = O(1/T).
\]
\end{proposition}

\begin{theorem}[PL contraction]
\label{thm:pl}
If $f$ satisfies a PL inequality with constant $\mu$ and the residual curvature outside the top-$k$ subspace is bounded, then for suitable $\alpha$ and refresh interval $m$, the expected loss at refresh steps contracts by a factor $(1-\gamma)$ depending on $(\{\lambda_i\}, \eta, \mu, m)$.
\end{theorem}

\section{Additional empirical details}
\subsection{Sensitivity to damping and refresh}
We swept damping $\eta$ in $\{1e{-3},1e{-2},1e{-1}\}$ and refresh $m\in\{200,400,800\}$ on CIFAR-100/ResNet-18 (single seed, not plotted). Larger $\eta$ reduces variance but slows progress; $m=400$ balances HVP cost and freshness, matching the main setting. Adaptive refresh based on gradient norm change is a promising extension.

\subsection{Ablation on threshold choice}
The speedup holds across thresholds in $[0.6,1.0]$: CAO hits 0.9 at 16 epochs vs Adam at $\approx$31, and 1.0 at 14 vs 23 (mean over three seeds; numbers from logs). We keep 0.75 for alignment with Figure~\ref{fig:main-full}.

\subsection{Failure modes and sanity checks}
No divergence was observed on these tasks; CAO($k{=}0$) behaves like a scaled SGD baseline. At very small damping $\eta$, HVP noise can destabilize the estimate; we therefore fix constant $\eta$ and clip gradients.

\section{Appendix figures and cost table}
\begin{figure}[!t]
    \centering
    \includegraphics[width=0.92\columnwidth]{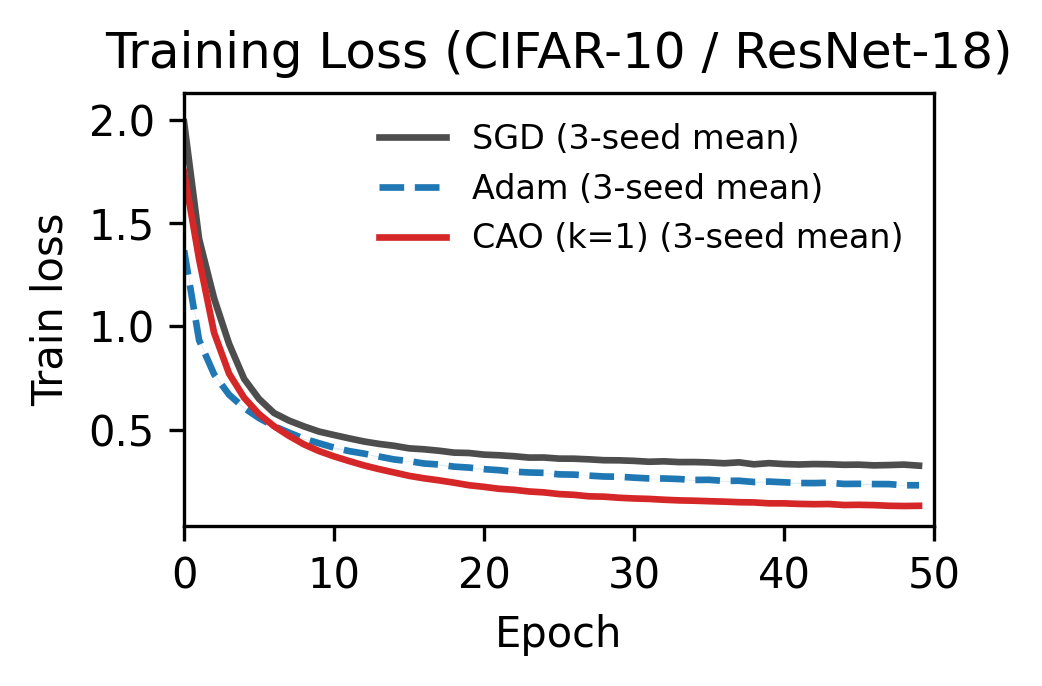}\vspace{-2mm}
    \caption{Appendix A: Train loss on CIFAR-10 / ResNet-18 (3 seeds), no threshold.}
    \label{fig:appendix-c10-r18}
\end{figure}

\begin{figure}[!t]
    \centering
    \includegraphics[width=0.92\columnwidth]{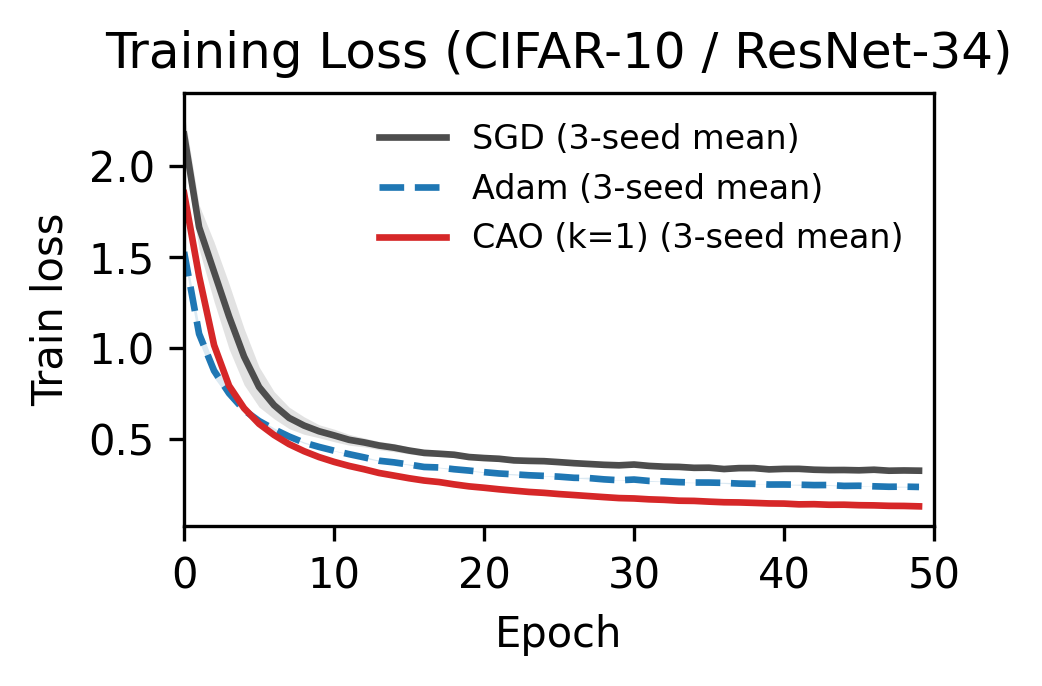}\vspace{-2mm}
    \caption{Appendix B: Train loss on CIFAR-10 / ResNet-34 (3 seeds), no threshold.}
    \label{fig:appendix-c10-r34}
\end{figure}

\begin{figure}[!t]
    \centering
    \includegraphics[width=0.92\columnwidth]{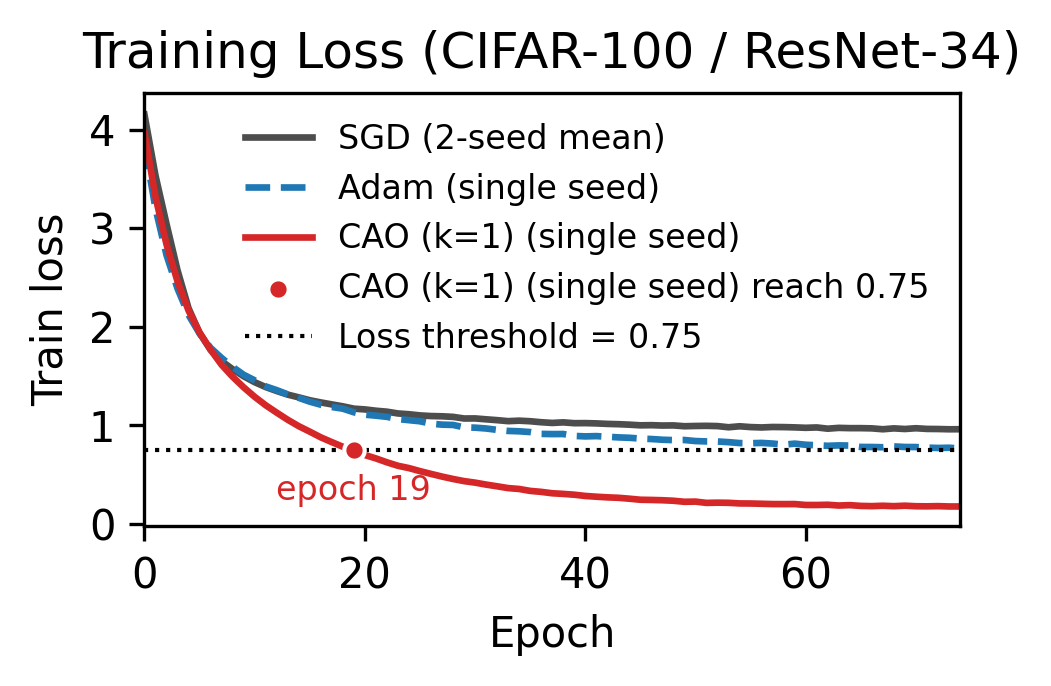}\vspace{-2mm}
    \caption{Appendix C: Train loss on CIFAR-100 / ResNet-34 (single seed).}
    \label{fig:appendix-c100-r34}
\end{figure}

\begin{figure}[!t]
    \centering
    \includegraphics[width=0.92\columnwidth]{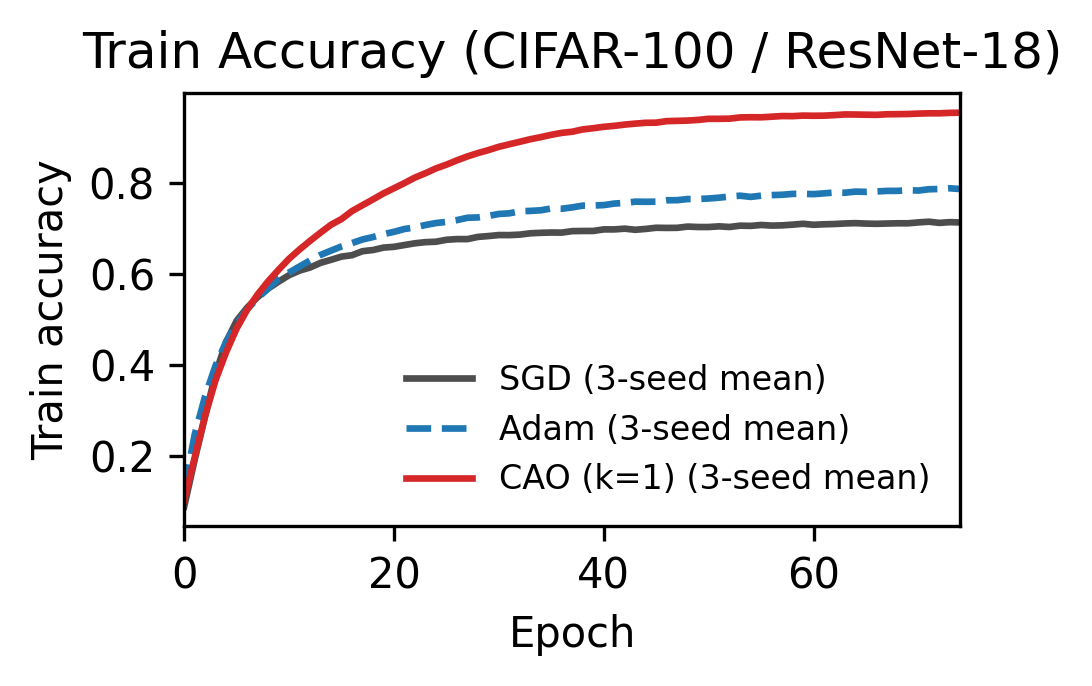}\vspace{-2mm}
    \caption{Appendix D: Train accuracy on CIFAR-100 / ResNet-18 (3 seeds).}
    \label{fig:appendix-acc}
\end{figure}

\begin{table}[!t]
    \centering
    \scriptsize
    \caption{Per-epoch wall-clock and peak memory; $\dagger$ single-seed.}
    \label{tab:cost}
    \setlength{\tabcolsep}{3pt}
    \resizebox{\columnwidth}{!}{
    \begin{tabular}{lccc}
    \toprule
    Setting & SGD & Adam & CAO (k=1) \\
    \midrule
    C10-R18 & 7.46 / 936 & 7.53 / 1003 & 8.78 / 2508 \\
    C10-R34 & 10.56 / 1593 & 10.76 / 1688 & 12.83 / 4405 \\
    C100-R18 & 7.40 / 1130 & 7.51 / 1143 & 8.77 / 2700 \\
    C100-R34 & 10.73$^{\dagger}$ / 1628$^{\dagger}$ & 10.89$^{\dagger}$ / 1537$^{\dagger}$ & 13.01$^{\dagger}$ / 4354$^{\dagger}$ \\
    \bottomrule
    \end{tabular}}
\end{table}

\end{document}